# A COUNTEREXAMPLE TO THE FORWARD RECURSION IN FUZZY CRITICAL PATH ANALYSIS UNDER DISCRETE FUZZY SETS

Matthew J. Liberatore[1]

[1]Department of Management and Operations, Villanova School of Business, Villanova University, Villanova, PA 19085

## ABSTRACT

*Fuzzy logic is an alternate approach for quantifying uncertainty relating to activity duration. The fuzzy version of the backward recursion has been shown to produce results that incorrectly amplify the level of uncertainty. However, the fuzzy version of the forward recursion has been widely proposed as an approach for determining the fuzzy set of critical path lengths. In this paper, the direct application of the extension principle leads to a proposition that must be satisfied in fuzzy critical path analysis. Using a counterexample it is demonstrated that the fuzzy forward recursion when discrete fuzzy sets are used to represent activity durations produces results that are not consistent with the theory presented. The problem is shown to be the application of the fuzzy maximum. Several methods presented in the literature are described and shown to provide results that are consistent with the extension principle.*

## 1. INTRODUCTION

CPM or the critical path method [11] has been successfully applied to plan and control projects that are organized as a set of inter-related activities. PERT or Program Evaluation and Review Technique [16] and Monte Carlo simulation apply probability analysis to address situations where there is uncertainty related to activity duration. PERT models uncertainty by collecting optimistic, most likely and pessimistic duration estimates of all activities and makes certain assumptions about the underlying probability distributions. Since the basic version of PERT tends to underestimate the expected minimum project duration [15]. Monte Carlo simulation is often preferred in practice when activity durations are uncertain.

However, the information required to estimate probabilities related to activity duration may not always be known. Fuzzy logic is an alternative approach for measuring uncertainty related to activity duration. Fuzzy logic measures imprecision or vagueness in estimation, and may be preferred to probability theory in those situations where past data concerning activity duration is either unavailable or not relevant, the definition of the activity itself is somewhat unclear, or the notion of the activity's completion is vague. Many authors including Chanas and colleagues have investigated the situation when activity duration can be described by fuzzy numbers [1], [2], [3], 5], [6], [7].

The dominant approach presented in the fuzzy critical path analysis literature is the fuzzy extension of the forward and backward recursions taken in the project network. This approach computes the earliest and latest start and finish times and slack, where the maximum, minimum, addition, and subtraction operators are replaced by their fuzzy counterparts. The application of the forward recursion with fuzzy activity times was first demonstrated in [3]. In a review of the fuzzy critical path analysis literature two approaches are described for applying the forward recursion [4]. They also indicate that the application of the backward recursion would cause a considerable increase in the range of uncertainty in the start and finish times that are calculated. These authors present a modification of the backward recursion that has been proposed to eliminate this disadvantage [12]. Some authors directly apply the backward recursion, while other authors have proposed different approaches for modifying the backward pass when the activity times are fuzzy [18], [19], [20], [22]. The backward recursion was found not to compute the sets of the possible values of the latest starting times and floats of activities [22]. In a stream of research that uses the joint possibility distribution of activity durations, several authors [8], [9] have conducted preliminary work for computing fuzzy latest starting times and fuzzy floats, especially for series-parallel graphs. Polynomial algorithms for determining the intervals of the latest starting times in the general project network are presented in [22].

Unlike the fuzzy backward recursion, the use of the fuzzy forward recursion is generally accepted in the literature as providing correct results. As mentioned by several authors [3], [10], [18], [19], the forward recursion is correct on problems involving fuzzy intervals. The purpose of this paper is to prove that the fuzzy forward recursion when discrete fuzzy sets are used to represent activity durations provides results that are not consistent with the direct application of the extension principle to the critical path problem.

In section 2, we present some background on fuzzy sets and the supporting concepts needed in the remainder of the paper. In section 3 we provide some brief background on project network analysis under the assumption that activity durations are certain. Section 4 presents an important result for fuzzy critical path analysis based on the extension principle. In section 5 the counterexample to the application of the fuzzy forward recursion is presented and the reason the results are not consistent with the extension principle is shown. Section 6 discusses two proposed approaches from the literature that present results consistent with the extension principle. Conclusions are offered in section 7.

## 2. FUZZY SETS

Following [17] we give the following definitions of a fuzzy set, fuzzy quantity and the support set:

**Definition 1.** A fuzzy set M is a subset of the universe $U$ that is characterized by a membership function $\mu_M : U \rightarrow [0,1]$ such that:

$\mu_M(x) = 0$, if $x$ certainly is not a member of $M$,

$\mu_M(x) = 1$, if $x$ certainly is a member of $M$,

$\mu_M(x) \in (0,1)$, if it is uncertain whether $x$ is a member of $M$, where $\mu_M(x)$ represents the degree to which $x$ is a member of $M$.

Now assume that activity duration is uncertain due to vagueness or imprecision and is a *fuzzy quantity* defined as follows:

**Definition 2.** A fuzzy quantity $M$ is a fuzzy subset of  with membership function $\mu_M: \rightarrow [0,1]$ such that

$$\sup(\mu_M(x) : x \in ) = 1 \tag{1a}$$

$$\exists x_1^M, x_2^M \in , x_1^M < x_2^M : \forall x \in , \\ x \notin (x_1^M, x_2^M) \Rightarrow \mu_M(x) = 0 \tag{1b}$$

**Definition 3.** The *support set* of $\mu_M$ is defined as $S_M = \{x \in \mid \mu_M(x) > 0\}$.

This definition can include discrete as well as continuous fuzzy sets, such as those that are triangular or trapezoidal in shape. If the support set $S_M$ is finite, condition (1a) above can be replaced with $\exists x_0 \in : \mu_M(x_0) = 1$. To simplify the presentation, we will assume this condition hereafter. Condition (1b) implies that the support set is bounded. We will also assume that the support set is discrete. Let $S_i$ be the support set of $a_i$, where $t_{i,k} \in S_i, k = 1, 2, ..., n_i$. That is, $a_i$ has $n_i$ possible non-zero discrete durations $t_{i,k}$ in its support set $S_i$.

As an example of a fuzzy activity duration, the vague statement that the duration of activity $a_i$ is "about six weeks" might be represented by the following fuzzy quantity $\tilde{t}_i$ whose membership function will be denoted as: $\mu_i(6) = .7, \mu_i(7) = 1.0, \mu_i(8) = .8, \mu_i(t_i) = 0$, otherwise. We can also denote this membership function as 6/0.7, 7/1.0, 8/0.8.

We also need to define fuzzy addition and the fuzzy maximum of fuzzy quantities. Let $M_1$ and $M_2$ be fuzzy quantities with membership functions $\mu_{M_1}$ and $\mu_{M_2}$, respectively. Following [17] and as a direct application of the extension principle:

**Definition 4. Fuzzy Addition:** The membership function for the fuzzy quantity $M_1 \oplus M_2$ is defined as:

$$\mu_{M_1 \oplus M_2}(z) = Max \quad Min \{\mu_{M_1}(x), \mu_{M_2}(y)\} \tag{2}$$

$$z = x + y$$

**Definition 5. Fuzzy Maximum:** The membership function for the fuzzy quantity max $(M_1, M_2)$ is defined as:

$$\mu_{\max(M_1, M_2)}(z) = \max_{z = \max(x, y)} \min\{\mu_{M_1}(x), \mu_{M_2}(y)\} \qquad (3)$$

One final definition is needed:

**Definition 6**: A *configuration* $\Omega$ is defined as a precise instantiation of the duration of all $a_i \varepsilon A$. In the next section we provide the necessary background on project network analysis.

## 3. BACKGROUND ON PROJECT NETWORKS AND THE CRITICAL PATH

Let $A = \{a_1, a_2, ..., a_N\}$ be the set of project activities. Let $B_i$, $i=1,2,...,N$, $B_i \subset A$, be defined so that the elements of $B_i$ are the immediate predecessors of $a_i$. We let $B = \{B_1, B_2, ..., B_N\}$ be the set of predecessor sets. Cycles of activities within the project network are not allowed. We assume that $a_1$ is the only element of $A$ such that $B_1 = \emptyset$. We also assume $a_N$ is the only element of $A$ such that $a_N \notin B_i \forall i \neq N$. If the project network does not have unique start and finish activities, we add dummy activities (which have zero duration) for this purpose. Taken together $A$ and $B$ define the network structure or graph $G$ of a project.

We initially assume that $t_i \in$ , $t_i \geq 0$, is the certain duration for $a_i$, where $T = \{t_1, t_2, ..., t_N\}$. A path $p$ from start to finish is a finite sequence of activities $\{a_{r_1}, a_{r_2}, ..., a_{r_s}\}$ where $a_{r_k} \in B_{r_{k+1}}, k = 1, 2, ..., s-1$, and $a_{r_1} = a_1$ and $a_{r_s} = a_n$. The length of the longest path from the start activity to the finish activity is the minimum project completion time. This path is called the *critical path*, the activities along it are called *critical path activities*, and the length of the path is denoted as $L_{CP}$. Given $G$ for each unique set of values $T$ we can compute $L_{CP}$. Therefore, there is a function $f(T|G)$: $\times \times ... \times \rightarrow$ that maps $T$ to $L_{CP}$, or $L_{CP} = f(T|G)$.

## 4. FUZZY CRITICAL PATH ANALYSIS

**Proposition 1: Fuzzy Critical Path Membership Function**

The membership function for the fuzzy set of critical path lengths can be determined as:

56

$$\mu_{CP}(L_{CP}) = \underset{\substack{t_1,t_2,...,t_N \\ f(t_1,t_2,...,t_N \mid G) = L_{CP}}}{Max} \underset{i \in \{1,2,...,N\}}{Min} \{\mu_i(t_i)\} \quad (4)$$

where $\mu_{CP}(L_{CP})$ is the membership function for the length of the critical path over the fuzzy subset *CP* in ; and for a graph *G* and activity durations $t_1, t_2, ..., t_N$, $f(t_1,t_2,...,t_N \mid G)$ determines the length of the critical path.

**Proof:** direct application of the extension principle of fuzzy logic [21] using maximum and minimum for the disjunction and conjunction operators, respectively.

Because Proposition 1 represents the direct generalization of critical path analysis from the crisp to the fuzzy domain, any proposed fuzzy critical path analysis approach should provide results that are consistent with it.

One approach for implementing Proposition 1 requires defining all possible configurations, determining the belief of each configuration as the minimum of the beliefs associated with all activity durations included in this configuration, and then determining the maximum belief of all configurations leading to each possible critical path length. The results form the fuzzy set of critical path lengths.

## 5. COUNTER EXAMPLE TO THE FUZZY FORWARD RECURSION

In standard CPM the earliest start ($ES_i$) and earliest finish ($EF_i$) of an activity are defined as:

$$ES_i = \max_{j \in B_i} EF_j \quad (5)$$

where we set $ES_0 = 0$, and
$$EF_i = ES_i + t_i \quad (6)$$

As defined in section 3 activity *N* does not have any successors, and so $EF_N$ is the length of the critical path.

Assuming $\tilde{t}_i$ is a fuzzy quantity representing the duration of $a_i$, the fuzzy earliest start $\widetilde{ES}_i$ and fuzzy earliest finish $\widetilde{EF}_i$ are defined as follows:

$$\widetilde{ES}_i = \max_{j \in B_i} \widetilde{EF}_j \quad (7)$$

$$EF_i = ES_i \oplus \tilde{t}_i \tag{8}$$

where max and $\oplus$ represent the extended fuzzy maximum and fuzzy addition, respectively, as given in definitions 4 and 5.

**Proposition 2:** The fuzzy forward recursion does not always yield membership functions for the fuzzy set of critical path lengths that are consistent with Proposition 1.

**Proof:** By counterexample using the data from a simple series – parallel graph (these graphs are defined in [8] as shown in Figure 1. In this example there 2*3*3*1 = 18 configurations as shown in Table 1. For each configuration, we determine the lengths of all paths, identify the critical path, and following Proposition 1 determine the configuration belief as the minimum of the beliefs of the possible activity durations. We combine the configurations by taking the maximum of the beliefs over all configurations having the same critical path length. From Table 1 the fuzzy set of critical path lengths is shown to be:

$$\mu_{CP}(L_{CP}) = 6/0.1, 7/0.2, 8/0.5, 9/0.2, 10/0.5, 11/0.1, 12/1 \tag{9}$$

The fuzzy forward recursion is applied using equations (5) and (6), with fuzzy addition and fuzzy maximum as defined in equations (2) and (3), respectively. The results are as follows:

$$ES_1 = 0/1 \tag{10}$$

$$EF_1 = 0/1 \oplus (3/0.5, 5/1) = 3/0.5, 5/1 \tag{11}$$

$$ES_2 = EF_1 = 3/0.5, 5/1 \tag{12}$$

$$EF_2 = (3/0.5, 5/1) \oplus (3/0.2, 5/0.5, 7/1) = 6/0.2, 8/0.5, 10/0.5, 12/1 \tag{13}$$

$$ES_3 = EF_1 = 3/0.5, 5/1 \tag{14}$$

$$EF_3 = (3/0.5, 5/1) \oplus (2/0.1, 4/1, 6/0.1) = 5/0.1, 7/0.5, 9/1, 11/0.1 \tag{15}$$

$$\begin{aligned} ES_4 &= \max \{ EF_2, EF_3 \} \\ &= \max \{(6/0.2, 8/0.5, 10/0.5, 12/1), (5/0.1, 7/0.5, 9/1, 11/0.1)\} \\ &= 6/0.1, 7/0.2, 8/0.5, 9/0.5, 10/0.5, 11/0.1, 12/1 \end{aligned} \tag{16}$$

$$\begin{aligned} EF_4 &= (6/0.1, 7/0.2, 8/0.5, 9/0.5, 10/0.5, 11/0.1, 12/1) \oplus 0/1 \\ &= 6/0.1, 7/0.2, 8/0.5, 9/0.5, 10/0.5, 11/0.1, 12/1 \end{aligned} \tag{17}$$

where $EF_4$ is by definition $\mu_{CP}(L_{CP})$, the fuzzy set of critical path lengths.

A comparison of equations (9) and (17) show that the range of possible critical path lengths is the same, but that the belief of the possible critical path length of 9 is 0.2 in equation (9) and 0.5 in equation (17). This difference proves the assertion.

**Discussion:** The forward algorithm is generally regarded as providing correct results in the literature, with no distinction made for activity durations represented as discrete fuzzy sets. The difference in the belief for the possible critical path length of 9 from Example 1 results from incorrect comparisons made in the computation of the fuzzy maximum of the earliest finishes. Specifically, the fuzzy maximum incorrectly compares earliest finish values that are based on activity durations from different configurations. These configurations differ in terms of the possible values of the activity that is the immediate predecessor of those activities whose earliest finishes are being compared.

For example, in equation (16) the possible time 8/0.5 from $EF_2$ is compared with the possible time 9/1 from $EF_3$ to yield 9/0.5. Folding back, in equation (13) the possible duration 8/0.5 results from the addition of 3/0.5 from $EF_1$ and 5/0.5 from the possible duration of $a_2$. Since the $ES_1$ is 0, 3/0.5 is the possible duration of $a_1$. In equation (15) the possible duration 9/1 results from the addition of 5/1 from $EF_1$ and 4/1 from the possible duration of $a_3$. Since the $ES_1$ is 0, 5/1 is the possible duration of $a_1$. That is, different possible durations of $a_1$ are used to obtain the possible time 8/0.5 from $EF_2$ and the possible time 9/1 from $EF_3$. Thus, the fuzzy maximum incorrectly compares earliest finishes that are based on activity durations from different configurations. In the example above, if the possible duration of $a_1$ were 3/0.2 instead of 3/0.5, then the fuzzy forward recursion would have produced the correct result, simply because of the specific value of a belief. However, since the fuzzy maximum is used repeatedly in the application of the fuzzy forward recursion, errors are likely to occur in many, if not most, problems.

# 6.PROPOSED CRITICAL PATH METHODS THAT ARE CONSISTENT WITH THE EXTENSION PRINCIPLE

This author has proposed several alternative approaches for computing the fuzzy set of critical path lengths that are consistent with the extension principle and Proposition 2. The first approach [13] is based on the idea that rather than enumerate all possible configurations, generate and then analyze a small number of configurations to achieve convergence to the membership function of the fuzzy set of critical path lengths. This method uses pseudo-random numbers to select possible durations for each activity to form configurations. Either the standard forward pass or path enumeration can be applied to determine the critical path length of a configuration. Following Proposition 1, the belief associated with the possible critical path length of a given configuration is the minimum of the beliefs of the possible activity durations in the configuration. Since several configurations may lead to the same possible critical path length, again following Proposition 1 and the extension principle, the belief of a possible critical path length is the maximum of the beliefs associated with all possible critical path lengths of the same duration. Additional configurations are analyzed until the percentage change in AREA (sum of the products of all possible critical path lengths and their beliefs) is less than some tolerance. The proposed method is evaluated using a series of test problems from the literature.

The second approach [14] begins by using a standard forward-backward pass algorithm to compute the minimum possible critical path length, using the smallest possible duration for each activity. A similar approach determines the maximum possible critical path length. Following Proposition 1 a mathematical programming problem is formulated, whose objective is to determine the belief associated with a possible critical path length. This mathematical program is run for all possible critical path lengths, ranging between the possible minimum and maximum lengths. In this way, the membership function for the fuzzy set of critical path lengths is constructed. The proposed method is evaluated using a series of test problems from the literature.

## 7. CONCLUSIONS

In this paper, the direct application of the extension principle leads to a proposition that must be satisfied in fuzzy critical path analysis. Using a counterexample where discrete fuzzy sets are used to represent activity durations, it is demonstrated that the fuzzy forward recursion produces results that are not consistent with the theory presented. The problem is shown to be the application of the fuzzy maximum, since it compares earliest finishes that are based on activity durations from different configurations. Several methods offered in the literature that provide results consistent with the extension principle are described. The development of more efficient algorithms to compute the fuzzy set of critical path lengths is an area of further research.

Figure 1. Example network with fuzzy activity durations

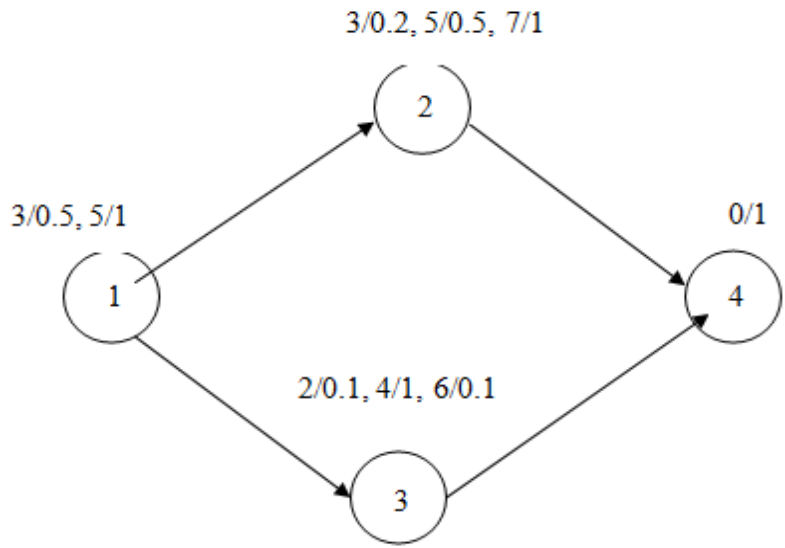

**Table 1: Applying proposition 1 to the example**

| Activity 1 | | Activity 2 | | Activity 3 | | P 1 - 2 - 4 | P 1 - 3 - 4 | Critical Path | | CP fuzzy set | |
|---|---|---|---|---|---|---|---|---|---|---|---|
| Duration | Belief | Duration | Belief | Duration | Belief | Duration | Duration | Duration | Belief | Duration | Belief |
| 3 | 0.5 | 3 | 0.2 | 2 | 0.1 | 6 | 5 | 6 | 0.1 | 6 | 0.1 |
| 3 | 0.5 | 3 | 0.2 | 4 | 1 | 6 | 7 | 7 | 0.2 | 7 | 0.2 |
| 3 | 0.5 | 3 | 0.2 | ´ | 0.1 | 6 | 9 | 9 | 0.1 | 8 | 0.5 |
| 3 | 0.5 | 5 | 0.5 | | 0.1 | 8 | 5 | 8 | 0.1 | 9 | 0.2 |
| 3 | 0.5 | 5 | 0.5 | 4 | 1 | 8 | 7 | 8 | 0.5 | 10 | 0.5 |
| 3 | 0.5 | 5 | 0.5 | 6 | 0.1 | 8 | 9 | 9 | 0.1 | 11 | 0.1 |
| 3 | 0.5 | 7 | 1 | 2 | 0.1 | 10 | 5 | 10 | 0.1 | 12 | 1 |
| 3 | 0.5 | 7 | 1 | 4 | 1 | 10 | 7 | 10 | 0.5 | | |
| 3 | 0.5 | 7 | 1 | 6 | 0.1 | 10 | 9 | 10 | 0.1 | | |
| 5 | 1 | 3 | 0.2 | 2 | 0.1 | 8 | 7 | 8 | 0.1 | | |
| 5 | 1 | 3 | 0.2 | 4 | 1 | 8 | 9 | 9 | 0.2 | | |
| 5 | 1 | 3 | 0.2 | 6 | 0.1 | 8 | 11 | 11 | 0.1 | | |
| 5 | 1 | 5 | 0.5 | 2 | 0.1 | 10 | 7 | 10 | 0.1 | | |
| 5 | 1 | 5 | 0.5 | 4 | 1 | 10 | 9 | 10 | 0.5 | | |
| 5 | 1 | 5 | 0.5 | 6 | 0.1 | 10 | 11 | 11 | 0.1 | | |
| 5 | 1 | 7 | 1 | 2 | 0.1 | 12 | 7 | 12 | 0.1 | | |
| 5 | 1 | 7 | 1 | 4 | 1 | 12 | 9 | 12 | 1 | | |
| 5 | 1 | 7 | 1 | 6 | 0.1 | 12 | 11 | 12 | 0.1 | | |

*activity 4 is a dummy whose duration is certainly 0